%% file: manuscript.tex
\documentclass{IEEEtran}
\usepackage{framed,multirow}
\usepackage{cite}
\usepackage{amsmath,amssymb,amsfonts}
\usepackage{algorithmic}
\usepackage{graphicx}
\usepackage{textcomp}
\usepackage{multirow}
\usepackage{comment}
\usepackage{tabu}
\usepackage{makecell}
\usepackage[colorlinks=true]{hyperref}
\usepackage{graphicx}
\usepackage[table,xcdraw]{xcolor}
\usepackage[normalem]{ulem}
\useunder{\uline}{\ul}{}

\def\etal{\emph{et al.}}
\def\ie{\emph{i.e.}}
\def\eg{\emph{e.g.}}

\def\cf{\emph{c.f.}}

\begin{document}
\title{Two-stage Cytopathological Image Synthesis for Augmenting Cervical Abnormality Screening}
\author{
Zhenrong Shen$^\dagger$, Manman Fei$^\dagger$, Xin Wang, Jiangdong Cai, Sheng Wang, Lichi Zhang, and Qian Wang 

\thanks{$^\dagger$ Equally contributed to this work.}
\thanks{Zhenrong Shen, Manman Fei, Xin Wang, Sheng Wang, and Lichi Zhang are with the School of Biomedical Engineering, Shanghai Jiao Tong University, Shanghai, China. (e-mail: \{zhenrongshen, feimanman, wangxin1007, wsheng, lichizhang\}@sjtu.edu.cn).}
\thanks{Jiangdong Cai and Qian Wang are with the School of Biomedical Engineering \& State Key Laboratory of Advanced Medical Materials and Devices, ShanghaiTech University, Shanghai, China. (e-mail: \{caijd2022, qianwang\}@shanghaitech.edu.cn).}
}

\maketitle

\begin{abstract}
Automatic thin-prep cytologic test (TCT) screening can assist pathologists in finding cervical abnormality towards accurate and efficient cervical cancer diagnosis. 
Current automatic TCT screening systems mostly involve abnormal cervical cell detection, which generally requires large-scale and diverse training data with high-quality annotations to achieve promising performance.
Pathological image synthesis is naturally raised to minimize the efforts in data collection and annotation.
However, it is challenging to generate realistic large-size cytopathological images while simultaneously synthesizing visually plausible appearances for small-size abnormal cervical cells.
In this paper, we propose a two-stage image synthesis framework to create synthetic data for augmenting cervical abnormality screening.
In the first \textit{Global Image Generation} stage, a \textit{Normal Image Generator} is designed to generate cytopathological images full of normal cervical cells.
In the second \textit{Local Cell Editing} stage, normal cells are randomly selected from the generated images and then are converted to different types of abnormal cells using the proposed \textit{Abnormal Cell Synthesizer}.
Both \textit{Normal Image Generator} and \textit{Abnormal Cell Synthesizer} are built upon Stable Diffusion, a pre-trained foundation model for image synthesis, via parameter-efficient fine-tuning methods for customizing cytopathological image contents and extending spatial layout controllability, respectively.
Our experiments demonstrate the synthetic image quality, diversity, and controllability of the proposed synthesis framework, and validate its data augmentation effectiveness in enhancing the performance of abnormal cervical cell detection.
\end{abstract}


\section{Introduction}
\begin{figure*}[ht]
    \centering
    \includegraphics[width=\textwidth]{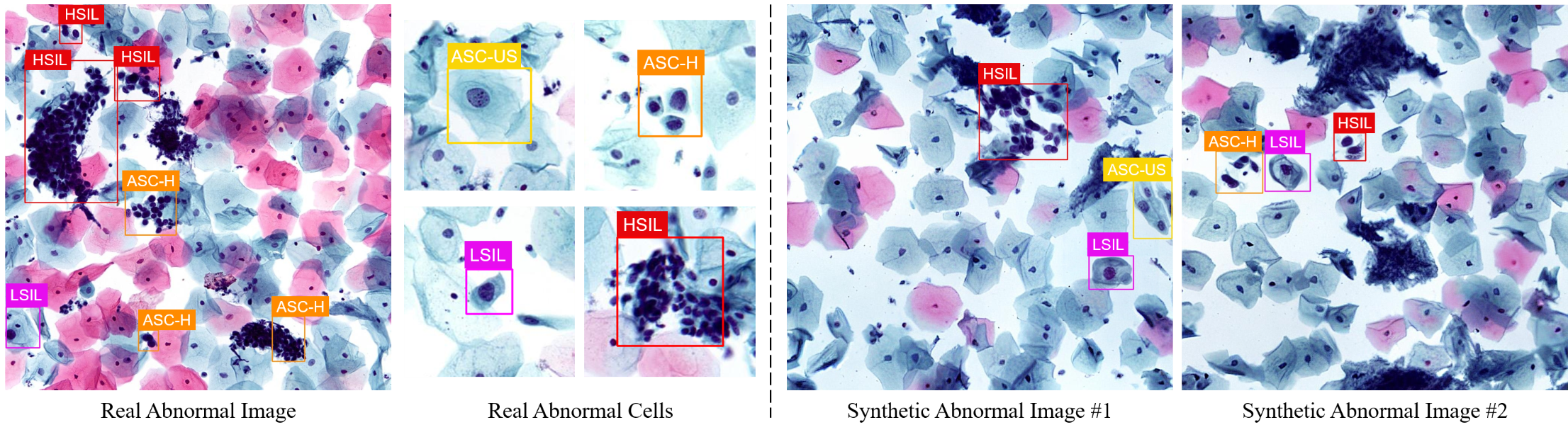}
    \caption{A real abnormal cervical cytopathological image and four real abnormal cell patches including ASC-US, ASC-H, LSIL, and HSIL are displayed on the left. Two examples of synthetic abnormal cervical cytopathological images produced by our proposed framework are shown on the right.}
    \label{fig:example}
\end{figure*}

\IEEEPARstart{C}{ervical} cancer is one of the serious healthcare threats that accounts for 6.6\% of the total cancer deaths in females worldwide~\cite{sung2021global}.
Early cytopathology screening is highly effective for preventing such malignancy~\cite{cramer1974role}, while thin-prep cytologic test (TCT)~\cite{siebers2009comparison} is a widely used technique for this purpose around the world.
According to the Bethesda system~\cite{nayar2015bethesda}, pathologists report the TCT specimen as negative for intraepithelial malignancy (NILM) when no cytological abnormalities are found.
Meanwhile, abnormal squamous cells discovered in TCT screening can be categorized into four cell types that are manifestations of cervical abnormality of varying degrees~\cite{davey1994atypical}, including 
atypical squamous cells of undetermined significance (ASC-US),
atypical squamous cells that cannot exclude a high-grade squamous intraepithelial lesion (ASC-H),
low-grade squamous intraepithelial lesion (LSIL),  
and high-grade squamous intraepithelial lesion (HSIL).
Fig.~\ref{fig:example} demonstrates their morphological differences.
By observing cellular features (\eg, nucleus-cytoplasm ratio) and judging cell types, pathologists can provide diagnoses that are crucial to the clinical management of cervical abnormality.

In traditional TCT screening, it is time-consuming, tedious, and usually subjective for human experts to identify abnormal cells within a gigapixel whole slide image (WSI). 
The large population of TCT specimens versus the limited number of well-trained pathologists further creates an overwhelming workload, especially in regions with limited healthcare resources.
Hence, computer-assisted TCT screening is highly desired to assist pathologists in interpreting TCT efficiently and accurately.
In recent studies, several methods apply deep learning in automatic WSI analysis~\cite{xiang2020novel,zhou2021hierarchical,cao2021novel}, where they usually leverage object detection algorithms~\cite{zou2023object} to extract suspicious abnormal cells. 
While the detectors generally rely on large amounts of annotated training data (\ie, with bounding boxes for all abnormal cells) to achieve satisfactory performances, it is difficult to collect sufficient and high-quality training data due to the expertise required for WSI annotation as well as the sensitivity of patient privacy~\cite{shen2021nodule,shen2023image,zhao2024stbi}.

Medical image synthesis has emerged as a promising solution for addressing the data scarcity problem by generating ground truths for training. 
Nevertheless, the task of cervical cytopathological image synthesis mainly faces three challenges.
\textit{First}, there are dozens, if not hundreds, of cervical cells in a cytopathological image. 
Yet each cell only occupies a small fraction of the image space.
The complex spatial relationships (\eg, isolated, squeezed, and overlapped) of individual cells greatly enhance the diversity of the cell appearance (\eg, color, morphology, and texture). 
It is difficult to render fine-grained details within small-size abnormal cells when generating cytopathological images of much larger sizes.
\textit{Second}, in order to effectively train the cell detector, it is imperative to generate abnormal cells with diverse locations and cell types for data augmentation. 
The complex spatial distributions of cells in cytopathological images place high demands on the flexibility of controlling cell locations. 
Meanwhile, the differences between cell types can be subtle and are mainly related to nuanced cellular attributes, thus expecting fine granularity in modulating cell types in the synthesis.
\textit{Third}, the conflict between the scarcity of available images and the necessity for abundant data to train reliable generative models is also a common challenge to medical image synthesis.

Recently, diffusion models have gained enormous popularity due to their remarkable capabilities to generate high-quality images~\cite{ho2020denoising,song2020score}.
Learning from large-scale data, text-to-image (T2I) diffusion models, especially Stable Diffusion~\cite{rombach2022high}, have demonstrated unprecedented ability in creating a wide variety of image contents that closely resemble the input text descriptions.
These T2I diffusion models serve as foundation models for image synthesis, which can be adapted to a wide range of domains and tasks through transfer learning.
To unleash their potentials in a computationally efficient way, many studies have explored parameter-efficient fine-tuning (PEFT) methods in pursuit of model customization~\cite{gal2022image,ruiz2023dreambooth,kumari2023multi,Han_2023_ICCV} or additional controllability~\cite{li2023gligen,zhang2023adding,mou2023t2i}.
The impressive progress has inspired us to employ PEFT techniques to tailor T2I diffusion models for both few-shot synthesis and fine-grained conditioning in cervical cytopathological image synthesis.

\begin{figure*}[ht]
    \centering
    \includegraphics[width=\textwidth]{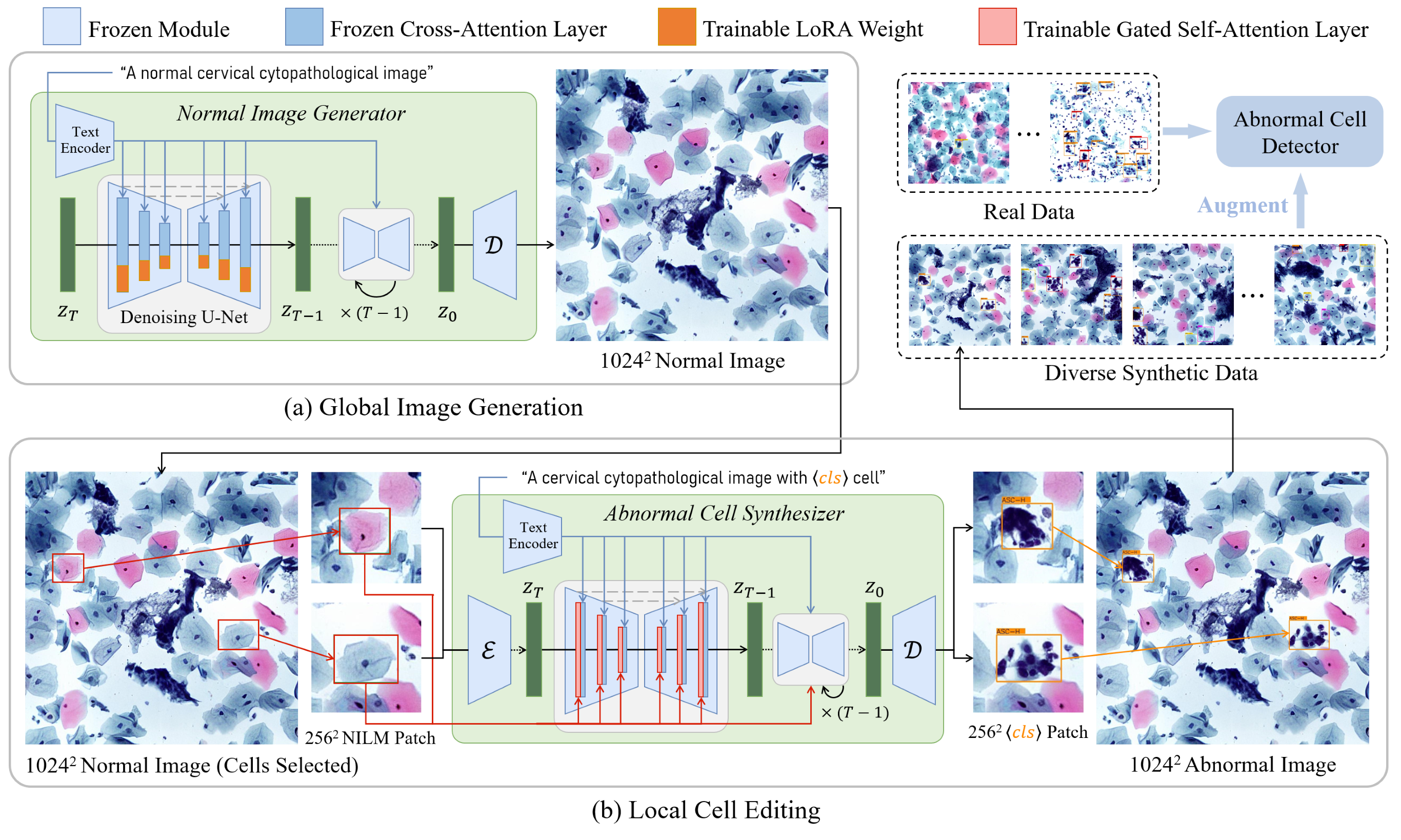}
    \caption{Our proposed cervical cytopathological image synthesis framework consists of two stages: (a) \textit{Global Image Generation}. Stable Diffusion is transferred to a \textit{Normal Image Generator} using Low-Rank Adaptation (LoRA) for creating high-resolution normal cytopathological images with NILM cells only. (b) \textit{Local Cell Editing}. A certain number of synthetic NILM cells are randomly selected and marked with bounding boxes using a simple in-house cell detector. Conditioned on the input texts and the bounding boxes, the proposed \textit{Abnormal Cell Synthesizer} can translate these selected cells to abnormal cells of user-defined types, thus obtaining diverse annotated abnormal cervical cytopathological images for data augmentation.}
    \label{fig:framework}
\end{figure*}

In this paper, we propose a novel cytopathological image synthesis framework to augment cervical abnormality screening by adapting T2I foundation models to the pathology domain. 
The proposed framework can synthesize diverse cytopathological images that incorporate cervical cells of various abnormality degrees as shown in Fig.~\ref{fig:example}.
The synthesis process naturally brings in a bounding box when placing an abnormal cell in the generated image. 
The synthetic data can thus be used to train a detector for suspicious abnormal cervical cells, which is essential to clinical usage. 
We empirically set the synthetic image size to be $1024\times1024$ with a microscope magnification of $\times20$, which strikes a balance between the number of cells in each image and the content of the image that a detector can effectively handle.
As illustrated in Fig.~\ref{fig:framework}, the overall framework is decomposed into two stages below:
\begin{itemize}
    \item \textit{\textbf{Global Image Generation}}. 
    We firstly leverage Low-Rank Adaptation (LoRA)~\cite{hu2021lora} to efficiently transfer the pre-trained Stable Diffusion to the \textit{Normal Image Generator} for customizing cytopathological image contents, which generates $1024\times1024$ cytopathological images full of NILM cells and other natural impurities.
    \item \textit{\textbf{Local Cell Editing}}.
    Based on a hypothesis that the neighboring effects between cervical cells are eliminated after oscillation and centrifugation in WSI preparation, we propose to synthesize abnormal cells by individually editing the cell types.
    Specifically, we employ a simple in-house cell detector to localize the synthetic NILM cells.
    After randomly selecting a certain number of them, we apply the \textit{Abnormal Cell Synthesizer}. 
    It can effectively extend the conditioning functionality of the pre-trained Stable Diffusion and translate the selected NILM cells into abnormal cells of user-defined types. 
\end{itemize}

In summary, the main contribution of our proposed framework is to address the aforementioned challenges as below:
\begin{itemize}
    \item We propose a new perspective on modeling cytopathological image synthesis as a two-stage framework at different image levels. This framework can faithfully model the global contexts of cell spatial relationships while meticulously rendering fine-grained cellular features of individual abnormal cells.
    \item The proposed framework can flexibly control the numbers, the locations, and the cell types of abnormal cells. Thus, large-scale and diverse synthetic abnormal cervical cytopathological images can be created to effectively augment cervical abnormality screening in various scenarios and even in previously unseen conditions.
    \item We incorporate PEFT techniques in building \textit{Normal Image Generator} and \textit{Abnormal Cell Synthesizer} upon Stable Diffusion, the state-of-the-art pre-trained foundation model for image synthesis, thus significantly alleviating the need for large-scale data in training reliable medical generative models.
\end{itemize}

\section{Related Works}
\subsection{Adapting Pre-trained Text-to-Image Models}
Trained on large-scale datasets of image-text pairs, text-to-image diffusion models (\eg, Stable Diffusion~\cite{rombach2022high}, Imagen~\cite{saharia2022photorealistic}, and DALL-E2~\cite{ramesh2022hierarchical}) have achieved state-of-the-art results in the field of text-guided image synthesis.
Despite their general capability of generating visually appealing image content according to the input text prompts, users often wish to harness their power for synthesizing specific concepts or more flexible controls.
To transfer pre-trained text-to-image models for specific concepts using only a few personalized images, recent works have focused on customizing models by fine-tuning the text embedding~\cite{gal2022image}, full weights~\cite{ruiz2023dreambooth}, cross-attention layers~\cite{kumari2023multi}, and singular values of the weight matrices~\cite{Han_2023_ICCV}.
To enable existing pre-trained text-to-image models to learn more task-specific input conditions, current studies endow them with new conditional input modalities by training additional self-attention layers~\cite{li2023gligen}, parallel networks~\cite{zhang2023adding}, and adapters~\cite{mou2023t2i}.
In this paper, we employ Low-Rank Adaptation (LoRA)~\cite{hu2021lora} to efficiently fine-tune Stable Diffusion for generating normal cervical cytopathological images. 
And we adopt the idea of GLIGEN~\cite{li2023gligen}, which can endow discrete conditioning (\ie, bounding box) capability to Stable Diffusion, to synthesize abnormal cells. 

\subsection{Pathological Image Synthesis}
A lot of studies have investigated pathological image synthesis methods for data augmentation on supervised deep learning algorithms of pathology analysis.
Existing literature has mainly explored histopathological image generation.
Hou~\etal~\cite{hou2019robust} proposed a hybrid synthesis pipeline that creates nucleus masks by predefined rules, initializes textures with real images, and utilizes a GAN model for further refinement.
AttributeGAN~\cite{ye2021multi} was designed for controllable histopathological image synthesis based on morphological cellular features.
HistoGAN~\cite{xue2021selective} produces histopathological images conditioned on class labels and involves a sample selection procedure for effective data augmentation.
Despite GAN-based methods~\cite{goodfellow2020generative}, recent works have also exploited diffusion models to synthesize histopathological images~\cite{ye2023synthetic}.
As for cytopathological image synthesis, CellGAN~\cite{shen2023cellgan} has been proposed to generate image patches of different cell types conditioned on class labels, but it fails to synthesize large images containing multiple cells.
In contrast, our current work is capable of generating synthetic data with larger image sizes, much more cells, and cell-level box annotations, thus enabling a broader range of data augmentation scenarios.

\section{Method}
\label{sec:method}
In this paper, we propose to disentangle the overall cervical cytopathological image synthesis into two stages, namely \textit{Global Image Generation} and \textit{Local Cell Editing}. 
As demonstrated in Fig.~\ref{fig:framework}, the first stage produces high-resolution normal cytopathological images, and the second stage synthesizes high-fidelity abnormal cell patches.
In the next, Section~\ref{subsec:stable_diffusion} reviews preliminary information about Stable Diffusion, which serves as the pre-trained foundation model in our work.
Section~\ref{subsec:image_generation} and Section~\ref{subsec:cell_editing} present \textit{Global Image Generation} and \textit{Local Cell Editing} in detail, respectively.

\subsection{Preliminary on Stable Diffusion}
\label{subsec:stable_diffusion}
Diffusion models achieve impressive results in image synthesis by decomposing the generation process into a series of denoising operations.
Among them,  Stable Diffusion, the successor of Latent Diffusion Model (LDM)~\cite{rombach2022high}, is one of the most powerful models publicly available.
To reduce the computational budgets, Stable Diffusion proceeds in a two-stage manner.
In the first stage, an autoencoder, which consists of an encoder $\mathcal{E}$ and a decoder $\mathcal{D}$, learns to obtain the latent representation $z_0=\mathcal{E}(x)$ of the input image $x$ with a downsampling factor of 8.
In the second stage, a denoising model $\epsilon_\theta$ is trained to progressively convert a Gaussian noise map $z_T$ into the desired latent representation $z_0$ through $T$ iterative steps.
Specifically, a forward process firstly gradually adds Gaussian noise $\epsilon_{t}$ to the latent representation $z_0$, forming a sequence of noisy samples $\{z_0,z_1,...,z_T\}$.
Starting from $z_T$, the denoising model $\epsilon_\theta$ generates a less noisy version $z_{t-1}$ of the input $z_t$ conditioned on a text prompt $c$ at each timestep $t$.
The training objective is simplified as below:

\begin{equation}
    \label{eq:objective}
    \mathcal{L}_{LDM}(\theta)=\mathbb{E}_{z,\epsilon\sim \mathcal{N}(\textbf{0},\textbf{I}),t,c}\left[{\left\|\epsilon_{\theta}(z_t,t,c)-\epsilon_{t}\right\|}_{2}^{2}\right] \\
\end{equation}

\noindent where $t$ is uniformly sampled from $\{1,...,T\}$.

\begin{figure}[t]
    \centering
    \includegraphics[width=\linewidth]{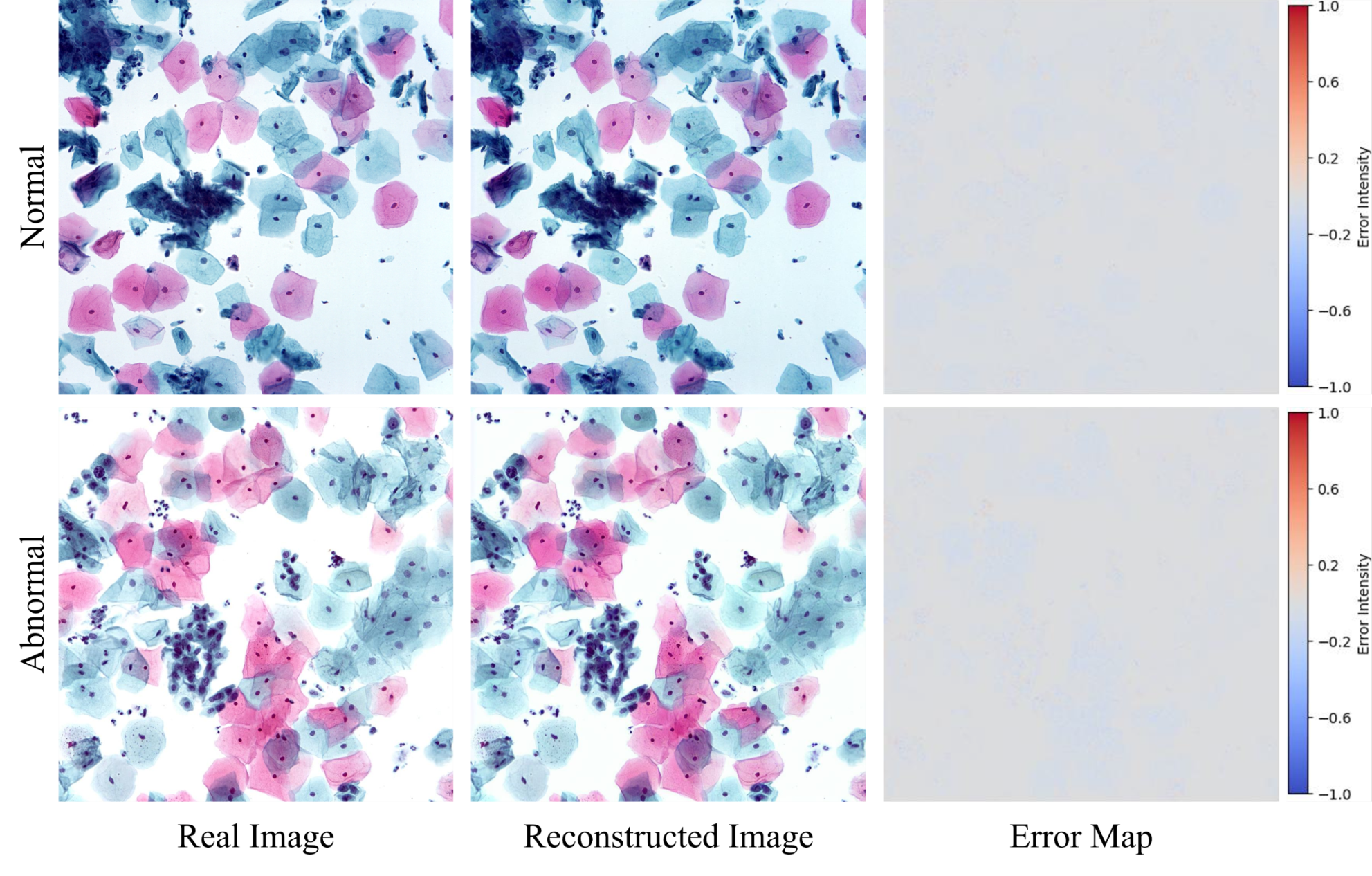}
    \caption{The pre-trained Stable Diffusion autoencoder can reconstruct both normal and abnormal cytopathological images well with negligible errors, suggesting that only the denoising U-Net needs to be fine-tuned.}
    \label{fig:vae_recon}
\end{figure}

In Stable Diffusion, the denoising model $\epsilon_\theta$ is implemented using a modified U-Net~\cite{ronneberger2015u}, which is sequentially formed by a series of ResNet~\cite{he2016deep} and Transformer~\cite{vaswani2017attention} blocks.
The timestep $t$ is mapped to a time embedding and injected into each ResNet block.
Meanwhile, the textual condition $c$ is embedded via the pre-trained CLIP text encoder~\cite{radford2021learning} and injected in the cross-attention layer within each Transformer block.
In each timestep $t$, the model takes in a noisy sample $z_t$ and produces a cleaner variant $z_{t-1}$ based on the text condition $c$, thus iteratively synthesizing an impressive image consistent with the text description.
In default, Stable Diffusion generates $512\times512$ images while its denoising U-Net processes $64\times64\times4$ latent maps.
As depicted in Fig.~\ref{fig:vae_recon}, the pre-trained autoencoder of Stable Diffusion is capable of faithfully reconstructing both normal and abnormal cytopathological images.
Thus, we only insert trainable modules into the U-Net in our proposed framework as shown in Fig.~\ref{fig:framework}. 

\subsection{Global Image Generation}
\label{subsec:image_generation}
To ensure the quality of synthetic images under the circumstance of limited training samples, we turn to Low-Rank Adaptation (LoRA)~\cite{hu2021lora} to effectively transfer Stable Diffusion to our \textit{Normal Image Generator} for generating normal cytopathological images in a computationally efficient way.
LoRA was first proposed to fine-tune large language models according to the hypothesis that the weight change of a pre-trained large model has high sparsity and a low intrinsic rank during fine-tuning.
It injects and optimizes trainable rank decomposition matrices into each layer of the Transformer architecture while keeping the original model weights frozen.
A recent study~\cite{zhu2023melo} has shown that LoRA can achieve comparable performance to fully fine-tuned models in medical image analysis tasks with a significant reduction in training data.

\begin{figure}[t]
    \centering
    \includegraphics[width=\linewidth]{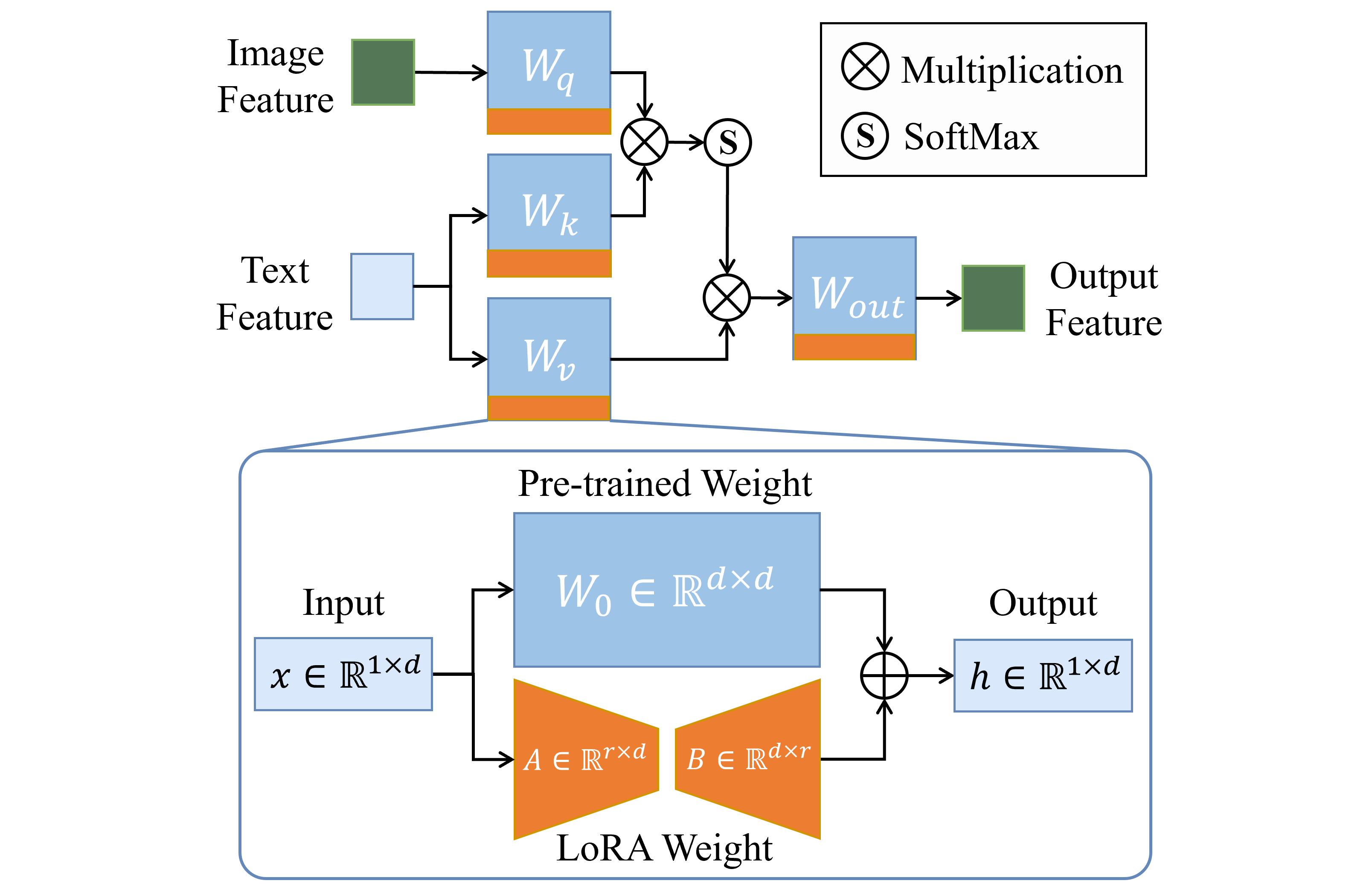}
    \caption{Low-Rank Adaptation (LoRA) is implemented in the pre-trained \textit{query}, \textit{key}, \textit{value}, and \textit{output} projection matrices of the pre-trained cross-attention modules. The LoRA weights learn to align image features with text embeddings for cytopathological image synthesis.}
    \label{fig:lora}
\end{figure}

\begin{figure*}[t]
    \centering
    \includegraphics[width=\textwidth]{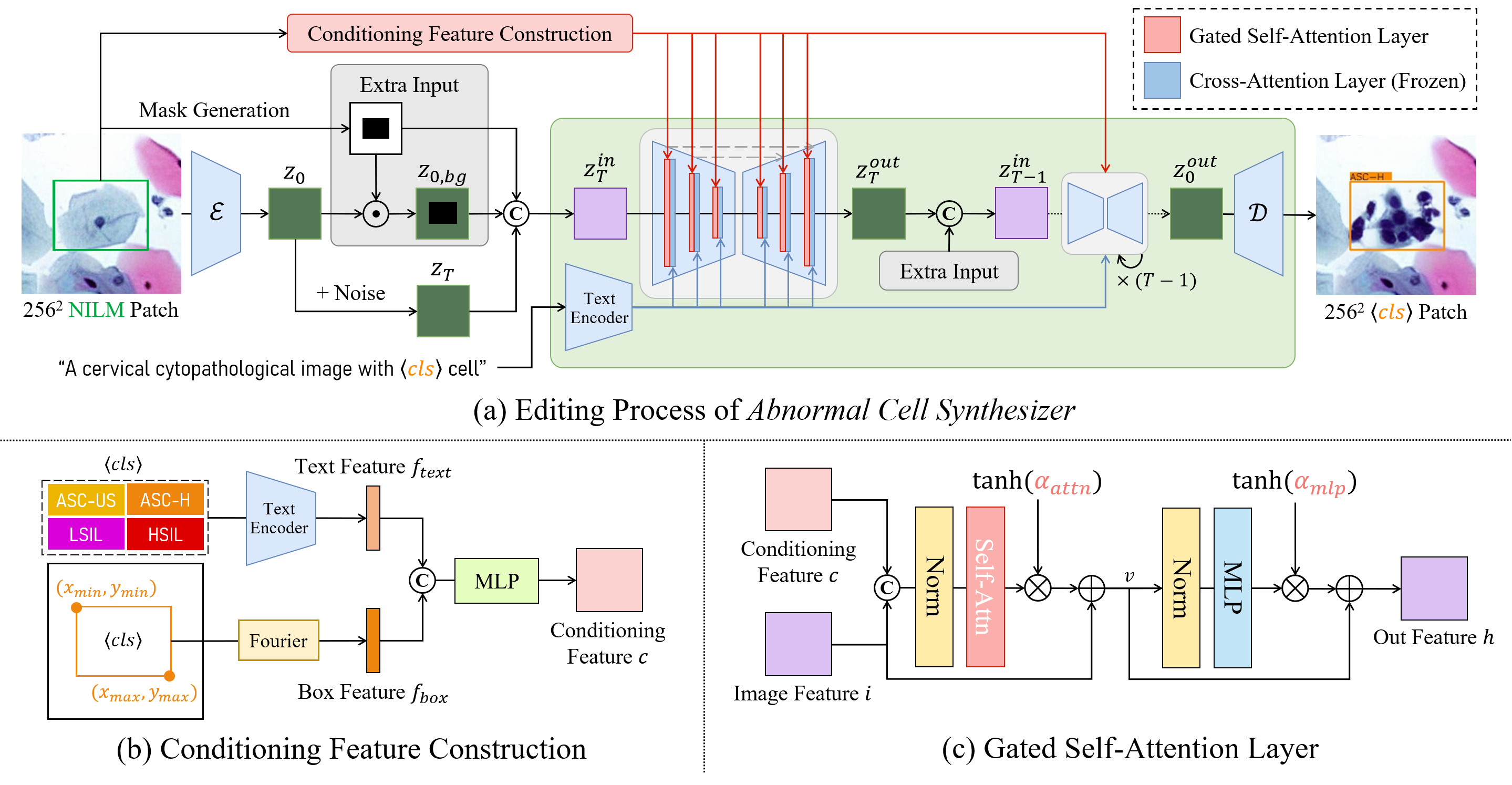}
    \caption{Detailed scheme of \textit{Local Cell Editing} stage includes (a) the editing process of \textit{Abnormal Cell Synthesizer}, (b) the construction process of conditioning feature, and (c) the module structure of gated self-attention layer. }
    \label{fig:gligen}
\end{figure*}

In Stable Diffusion, the text conditioning mechanism lies in the cross-attention layer where the CLIP-encoded text embedding is projected to the \textit{key} and \textit{value} representations and then interacts with the \textit{query} representation mapped from image features.
Therefore, we fix the input text prompt as \textit{``A normal cervical cytopathological image"}, and apply LoRA to the cross-attention modules in the denoising U-Net for aligning the image features with the predefined text description.
As shown in Fig.~\ref{fig:lora}, for the pre-trained \textit{query}, \textit{key}, \textit{value}, and \textit{output} projection matrices (denoted as $W_q$, $W_k$, $W_v$, and $W_{out}$) in each cross-attention layer, the injected LoRA weights constrain their updates with a low-rank decomposition by optimizing the objective in \eqref{eq:objective}, which is formulated as:

\begin{equation}
    h=W_{0}x+\Delta Wx=W_{0}x+BAx
\end{equation}

\noindent where $x\in\mathbb{R}^{1\times d}$ and $h\in\mathbb{R}^{1\times d}$ stand for the input and output features. $
W_{0}\in\mathbb{R}^{d\times d}$ and $\Delta W\in\mathbb{R}^{d\times d}$ denote the pre-trained weight and its change brought by LoRA.
$\Delta W=BA$ is decomposed into two low-rank matrices $B\in\mathbb{R}^{d\times r}$ and $A\in\mathbb{R}^{r\times d}$, and the rank $r$ is much smaller than the dimension $d$.
In our experiments, $A$ and $B$ are initialized with random Gaussian and zero numbers, respectively.
Given the difference between default and targeted image size, as well as the domain gap between natural and pathological images, we empirically set the rank $r$ to 8.
Needing only a few training images, \textit{Normal Image Generator} effectively learns the complicated spatial relationships and the varying appearances of NILM cells, thus capable of generating large and diverse sets of normal cytopathological images.

\subsection{Local Cell Editing}
\label{subsec:cell_editing}
In the first stage, the global contexts of cell spatial relationships are well simulated in the generated cytopathological images, but abnormal cells have not been synthesized yet.
According to the assumption that cervical cells are spatially independent of each other after oscillation and centrifugation in WSI preparation, we propose to explicitly edit the cell types of individual cells using a locally generative approach, which can produce realistic abnormal cells with box annotations for augmenting cervical abnormality screening, 
As depicted in Fig.~\ref{fig:framework}(b), we first use a simple in-house cell detector based on Faster R-CNN~\cite{ren2015faster} to label all NILM cells with bounding boxes.
Then a certain number of them are selected and cropped as multiple $256\times256$ image patches centered on their boxes.
The selection process can either be randomized or controlled by human input, depending on the desired outcome.
Next, we propose the \textit{Abnormal Cell Synthesizer}, which endows spatial layout controllability over pre-trained Stable Diffusion, to edit cell types conditioned on the designed prompts and the bounding boxes.
Finally, these edited patches are put to their original places, thus resulting in a cervical cytopathological image with multiple synthetic abnormal cells.

As illustrated in Fig.~\ref{fig:gligen}(a), the editing process of \textit{Abnormal Cell Synthesizer} is formulated in an inpainting scheme, thus producing abnormal cells consistent with the surrounding areas.
The encoder $\mathcal{E}$ converts a cropped $256\times256$ patch into a $32\times32\times4$ latent map $z_0$, which is further initialized to a Gaussian noise map $z_T$ using the diffusion forward process.
A binary mask is derived from the bounding box, downsampled to the same size of $z_0$, and applied to $z_0$ by element-wise multiplication, resulting in $z_{0,bg}$ to indicate the unedited background area at the feature level.
At each timestep $t$, both the binary mask and the masked latent map $z_{0,bg}$ are provided as extra inputs to the denoising U-Net.
For example, at the first timestep $T$, the initial Gaussian noise map $z_T$ is concatenated to the extra input, resulting in a 9-channel input tensor $z_{T}^{in}$. 
The denoising U-Net takes in $z_{T}^{in}$ and produces the output latent map $z_{T}^{out}$ with the same shape of $z_0$. 
$z_{T}^{out}$ is then concatenated to the extra input to form the input $z_{T-1}^{in}$ at the next timestep.
The rest iterations are done in the same manner so that the network can guarantee background preservation and improve the coherence of the edited result.
At last, the decoder $\mathcal{D}$ converts the final latent representation $z_{0}^{out}$ into the edited patch that contains the user-defined abnormal cell.

\begin{figure*}[ht]
    \centering
    \includegraphics[width=\textwidth]{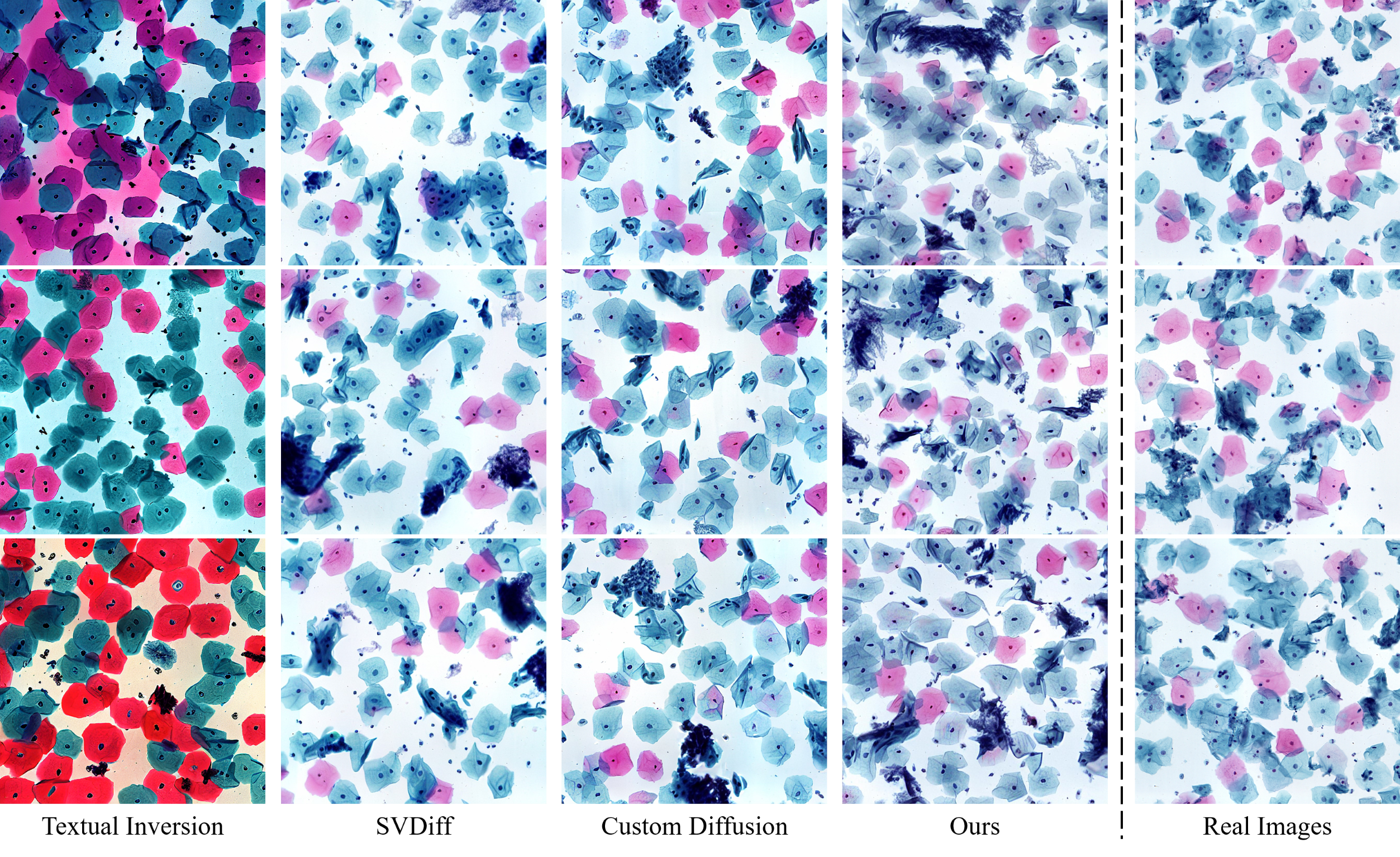}
    \caption{Qualitative comparison between different PEFT methods in transferring the pre-trained Stable Diffusion into \textit{Normal Image Generator}. Each method produces three examples in individual rows. Zoom in to check image details.}
    \label{fig:syn_compare}
\end{figure*}

In order to synthesize abnormal cells at the specified locations, we incorporate the bounding box as well as the cell type information into each denoising step through newly inserted gated self-attention layers.
As shown in Fig.~\ref{fig:gligen}(b), the bounding box is represented as [$x_{min}, y_{min}, x_{max}, y_{max}$] with its top-left and bottom-right coordinates while the targeted cell type is denoted by a text entity $\langle cls \rangle$. 
We obtain the box feature $f_{box}$ via Fourier embedding method~\cite{mildenhall2021nerf}, and then fuse it with the text feature $f_{text}$ encoded by the CLIP text encoder to produce a conditioning feature $c$:

\begin{equation}
    c=MLP([f_{text}, f_{box}])
\end{equation}

\noindent where a Multi-Layer Perceptron (MLP) concatenates $f_{text}$ and $f_{box}$ across the feature dimension as the input to construct the conditioning feature $c$. 

To enable spatial layout controllability in Stable Diffusion, we add trainable gated self-attention layers ahead of the cross-attention layers in every transformer block.
As displayed in Fig.~\ref{fig:gligen}(c), the gated self-attention layer is performed over the concatenation of image feature $i$ and conditioning feature $c$, and can be formulated as:

\begin{equation}
    v=i+{\rm tanh}(\alpha_{attn})\cdot {\rm SelfAttn}({\rm Norm}([i, c]))
\end{equation}

\begin{equation}
    h=v+{\rm tanh}(\alpha_{mlp})\cdot {\rm MLP}({\rm Norm}(v))
\end{equation}

\noindent where $v$ denotes the intermediate feature and $h$ denotes the output one.
Two learnable scalars $\alpha_{attn}$ and $\alpha_{mlp}$ are used for gating the outputs of self-attention module and MLP, respectively. 
They are both initialized with zero for training stability and then normalized to reasonable numerical ranges by hyperbolic tangent functions.
Such a gating mechanism automatically learns the optimal combination of original image features and additional conditional information.

Despite the conditioning process above, we place the text entity $\langle cls \rangle$ in the designed input text prompt \textit{``A cervical cytopathological image with $\langle cls \rangle$ cell"} to further guide image synthesis towards the user-defined cell type $\langle cls \rangle$ through the frozen cross-attention layers.
The original LDM objective in \eqref{eq:objective} is still used to adapt the pre-trained Stable Diffusion to \textit{Abnormal Cell Synthesizer}.

\section{Experimental Results}
\label{sec:results}

\input{tables/syn_compare}

\subsection{Dataset and Experimental Setup}
\label{sec:setup}
Our proposed two-stage cytopathological image synthesis framework is formulated in a two-stage manner, for which two different datasets and model settings are prepared as follows:

\begin{figure*}[ht]
    \centering
    \includegraphics[width=\textwidth]{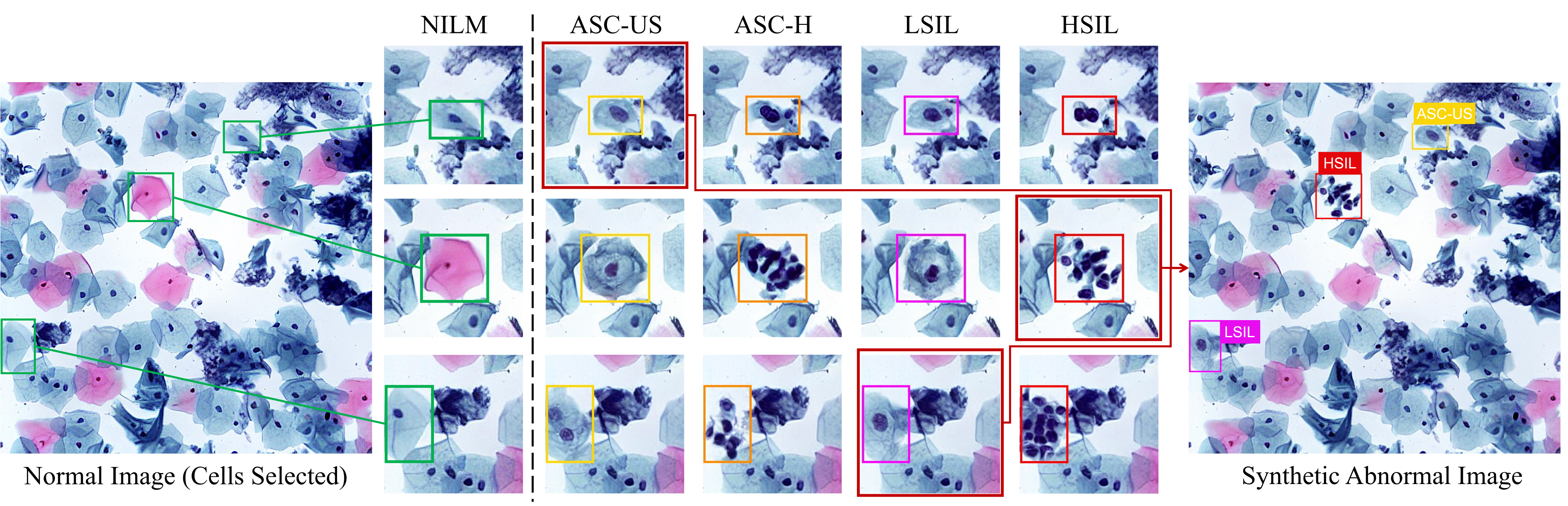}
    \caption{An example to illustrate the controls over the numbers, the locations, and the cell types of synthetic abnormal cells in \textit{Local Cell Editing} stage. Three NILM cells at different locations (marked by green boxes) are selected from a normal image, and then are translated into abnormal cells of different types using the proposed \textit{Abnormal Cell Synthesizer}. A random combination of edited abnormal cell patches (selected by red contours) is pasted back to the normal image, resulting in a synthetic abnormal cervical cytopathological image on the right of the figure.}
    \label{fig:syn_control}
\end{figure*}

\begin{itemize}
    \item \textit{\textbf{Global Image Generation}}.
    The training data for \textit{Normal Image Generator} contains 1,000 $1024\times 1024$ normal cervical cytopathological images.
    We use the learning rate of $1.0\times10^{-4}$, batch size of 1, and Adam optimizer~\cite{kingma2015adam} for training $5k$ iterations based on Stable Diffusion v1.5.
    The best model is selected for the following experiment.
    The Denoising Diffusion Implicit Model (DDIM) sampler~\cite{song2020denoising} with 25 steps and the classification-free guidance (CFG)~\cite{ho2021classifier} scale of 5.0 are implemented to generate images in the inference stage.
    
    \item \textit{\textbf{Local Cell Editing}}.
    We collect 3,651 $1024\times 1024$ cervical cytopathological images with abnormal cells from our collaborative hospitals.
    All the abnormal cells are well-annotated with bounding boxes by three experienced pathologists.
    We randomly select 3,000 images as the training data while the remaining 651 images are testing data for the detection task.
    From training images, we center-crop $256\times 256$ abnormal cell patches according to their box annotations, including 2,275 ASC-US, 2,480 LSIL, 1,638 ASC-H, and 422 HSIL cell patches in total.
    We use the learning rate of $1.0\times10^{-4}$, batch size of 4, and AdamW optimizer~\cite{loshchilov2018decoupled} to train \textit{Abnormal Cell Synthesizer} for $50k$ iterations based on Stable Diffusion.
    During inference, we use the DDIM sampler with 25 steps and the CFG scale of 5.0.
\end{itemize} 

To comprehensively evaluate our proposed framework, we conduct experiments from two aspects, including the synthetic image quality and the data augmentation effectiveness.
The implementation details as well as the results for each part of the experiments are presented in Section~\ref{subsec:eval_syn} and Section~\ref{subsec:eval_aug}, respectively.
All the experiments are conducted using an NVIDIA RTX A6000 GPU with PyTorch~\cite{paszke2019pytorch}.

\subsection{Evaluation of Synthetic Cervical Cytopathology}
\label{subsec:eval_syn}
\subsubsection{Comparison with Other PEFT Methods}
The synthetic image quality mainly depends on \textit{Global Image Generation} because the generated normal image (background) occupies the majority of the image space.
To comprehensively evaluate the results of \textit{Global Image Generation}, we compare our implementation with three different PEFT approaches in building \textit{Normal Image Generator}, including 
(a) Textual Inversion~\cite{gal2022image}, which only learns a new text embedding for a specific concept; 
(b) SVDiff~\cite{Han_2023_ICCV}, which fine-tunes the singular values of all the pre-trained weight matrices; 
(c) Custom Diffusion~\cite{kumari2023multi}, which updates the \textit{key} and \textit{value} projection matrices in the cross-attention layers and optimizes a learnable text embedding following Textual Inversion. 
All the methods are trained for $5k$ iterations, and the best models are selected.
Fr\'echet Inception Distance (FID)~\cite{heusel2017gans} is used to holistically measure the overall semantic realism of the generated normal images.
Specifically, we use the Inception V3~\cite{szegedy2016rethinking} model pre-trained on ImageNet~\cite{deng2009imagenet} and the ResNet-50 model pre-trained on the cervical cytopathology dataset~\cite{fei2023robust} as the feature extractors for a fair quantitative comparison from different aspects.
Each method generates 1,000 images for computing FID scores.

As shown in Fig.~\ref{fig:syn_compare}, Textual Inversion cannot preserve image color fidelity and brings forth varying color distortions across different generated samples.
The results of SVDiff tend to have cartoonish cytoplasmic textures rather than real-world photomicrographic appearances.
Custom Diffusion produces similar but inferior results to our method.
On the one hand, Custom Diffusion often yields black cell nuclei, whereas the cell nuclei generated by our method are dark blue which is more consistent with real images. 
On the other hand, Custom Diffusion only produces sharp cell boundaries as if each cell were individually pasted onto the same image plane. 
In contrast, our method can simultaneously generate clear and blurry cell boundaries, which is consistent with the visual effects of real microscope photography.
The quantitative comparison in Table~\ref{tab:syn_compare} further supports the superiority of our method in synthetic image quality.
Our method achieves higher FID scores than other methods for the feature extractors pre-trained on different image domains, suggesting that its synthetic data distribution is closer to the real one in the feature spaces of both natural and cytopathological images.
In summary, the proposed \textit{Normal Image Generator} based on LoRA is able to synthesize normal cervical cytopathological images with visually realistic cellular features and semantically plausible cell spatial relationships, thus providing a good base for further local editing of abnormal cells.

\input{tables/det_private}

\subsubsection{Controllability of Synthetic Abnormal Cells}
The diversity of the numbers, the locations, and the cell types of the synthetic abnormal cells can be controlled in \textit{Local Cell Editing}. 
For example in Fig.~\ref{fig:syn_control}, three NILM cells at specified locations are randomly selected from the detected cell candidates in a normal cytopathological image. 
\textit{Abnormal Cell Synthesizer} converts the selected cells into abnormal cells one by one, and allows users to control the level of cervical abnormality ranging from ASC-US to HSIL.
It is observed from Fig.~\ref{fig:syn_control} that the abnormal cellular features (\eg, enlarged hyperchromatic nuclei, shrunken cytoplasm, and crowding cell cluster) are well simulated for different cell types, which highlights the precise cellular feature modeling ability of \textit{Abnormal Cell Synthesizer}.
We utilize the feature extractor pre-trained on the cervical cytopathology dataset to compute FID scores for  $256\times256$ image patches of different abnormal cell types.
The quantitative results show that \textit{Abnormal Cell Synthesizer} can achieve 1.4136, 1.6463, 1.1856, and 1.3507 FID scores for ASC-US, ASC-H, LSIL, and HSIL cells, respectively, which indicates the high fidelity of synthetic cell morphology.
By converting NILM cells at any location to abnormal cells of varying malignant degrees, our proposed framework can provide diverse abnormal cervical cytopathological images to improve model robustness against various scenarios.

\subsection{Evaluation of Augmenting Abnormal Cell Detection}
\label{subsec:eval_aug}
To validate the data augmentation effectiveness using the proposed synthesis framework, different model architectures are applied for the task of cervical abnormal cell detection. The comparisons inlcude two-stage (\ie, Faster R-CNN~\cite{ren2015faster}), one-stage (\ie, RetinaNet~\cite{lin2017focal}), and transformer-based (\ie, DINO~\cite{zhang2022dino}) detectors. 
Following the experimental setup of \textit{Local Cell Editing} in Section~\ref{sec:setup}, we use the same training dataset of 3,000 real abnormal cervical cytopathological images to train each detector as the baseline model. 
Then we combine different kinds of synthetic data with these 3,000 real images for augmentation to train detectors. 
We use the remaining 651 real images as testing data for a fair comparison.
In this way, the data leakage issue is avoided in this experiment since \textit{Abnormal Cell Synthesizer} has not seen any testing samples during training, thereby ensuring the fairness of the data augmentation evaluation. 

\begin{figure}[t]
    \centering
    \includegraphics[width=\linewidth]{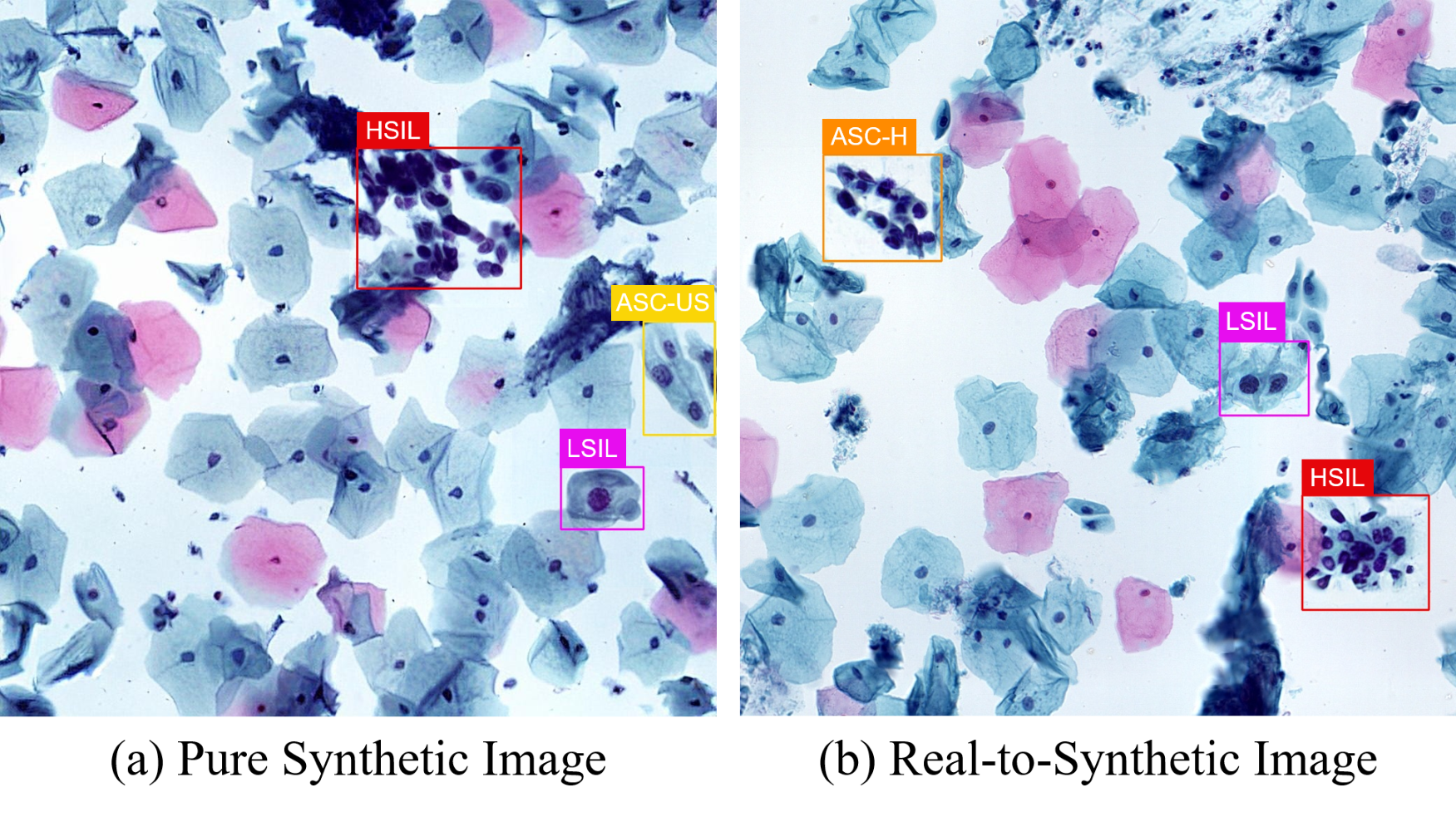}
    \caption{(a) Pure synthetic abnormal images are generated by fully executing the proposed two-stage synthesis framework. (b) Real-to-synthetic abnormal images are obtained by applying only \textit{Local Cell Editing} to edit NILM cells in each real normal cytopathological image.}
    \label{fig:s_vs_r2s}
\end{figure}

For Faster R-CNN and RetinaNet, we adopt SGD~\cite{robbins1951stochastic} to train models with the learning rate of $2.0\times10^{-4}$ and $1.0\times10^{-3}$, respectively.
As for DINO, we use the learning rate of $1.0\times10^{-4}$ and AdamW optimizer~\cite{loshchilov2018decoupled}.
All the detectors are trained with batch size 2 for 20 epochs.
% Random flip ($p=0.5$) is applied in all experiments.
We evaluate quantitative results using the COCO-style Average Precision (AP)~\cite{lin2014microsoft}, where AP is averaged over multiple Intersection over Union (IoU) thresholds from 0.50 to 0.95 with a step size of 0.05 for each abnormal cell type.
We also compute mean AP (mAP) by averaging APs over all the cell types, and individually evaluate mAPs at IoU thresholds of 0.5 and 0.75 (denoted by mAP$_{.5}$ and mAP$_{.75}$).
The following experiments illustrate the data augmentation effectiveness of our proposed framework in three aspects, including 1) the superiority over other synthesis methods, 2) the comparability with near-real data, and 3) the generalization to other image domains.

\input{tables/det_public}

\subsubsection{Comparison with Other Synthesis Methods}
The normal image quality generated by Custom Diffusion is close but inferior to LoRA used in our synthesis framework in both qualitative and quantitative comparisons (\cf, Section \ref{subsec:eval_syn}).
To further demonstrate their differences, we conduct a comparison between their data augmentation effectiveness.
Specifically, we first generate 5,000 normal images using each method, randomly select 1-8 NILM cells from each image, and edit them into abnormal cells.
Then we combine these 5,000 synthetic abnormal images and the original 3,000 real ones to form the training data for augmenting detectors.
The experimental results in Table~\ref{tab:det_private} show that adding synthetic data into training has effectively improved the performances of all the detectors compared with the baselines. 
For example, the mAP of DINO has increased from 15.9 to 17.8 using our synthetic data. 
In comparison to Custom Diffusion, the detectors trained with our synthetic data achieve higher mAP values and perform better in cell type identification. 
This demonstrates the effectiveness and superiority of our method in terms of data augmentation for the cell detection task.
Notably, all the detectors in Table~\ref{tab:det_private} achieve much lower AP values when recognizing HSIL cells.
We attribute this phenomenon to the intrinsic bias of our in-house dataset since Table~\ref{tab:det_public} shows different results when we conduct the experiment on another dataset.

\subsubsection{Pure Synthetic Data \textit{vs.} Real-to-Synthetic Data}
By fully executing our two-stage synthesis framework, one can theoretically synthesize an infinite number of abnormal cervical cytopathological images that do not exist in the real world. 
Meanwhile, real normal images can also be converted to abnormal images if only \textit{Local Image Editing} is used to directly edit real NILM cells into synthetic abnormal ones.
Intuitively, such real-to-synthetic images should display better visual effects than pure synthetic images, but their visual difference seems to be negligible as shown in Fig.~\ref{fig:s_vs_r2s}. 
So we are curious about their difference in data augmentation effectiveness when adding these two kinds of data into training. 

To answer this question, we collect another 5,000 normal cervical cytopathological images that are not used in training our synthesis framework in case of data leakage.
By employing \textit{Local Image Editing} to edit 1-8 randomly selected cells in each real normal image, we obtain 5,000 real-to-synthetic abnormal images and add them into training data for data augmentation.
As illustrated in Table~\ref{tab:det_private}, the detectors trained with pure synthetic data achieve comparable performance on par with those trained with real-to-synthetic data.
Among them, RetinaNet and DINO even reach higher mAP values, suggesting that the pure synthetic data from our proposed synthesis framework has the potential to substitute for real-to-synthetic data in augmenting downstream models. 
Furthermore, creating such real-to-synthetic data actually requires a substantial amount of real data that is inherently limited in both quantity and diversity. 
In contrast, our method can generate an arbitrary number and diverse range of data, making it particularly cost-effective and well-suited for large-scale applications where data collection and annotation can be costly and time-consuming.

\subsubsection{Generalization to Different Image Domains}
Although our synthetic data demonstrate high visual fidelity and effective model improvement, a shift in the image domain could still lead to a decrease in the data augmentation utility.
To demonstrate the robustness of our synthetic data across different image domains, we conduct an experiment on Comparison Detector database~\cite{liang2021comparison}, a publicly available dataset, to evaluate the data augmentation effectiveness.
The original Comparison Detector database contains 11 cell and bacteria categories. 
After filtering out categories except for ASC-US, ASC-H, LSIL, and HSIL, there are 4,651 training images and 505 test images in total.
The longest side of the image is resized to 1024 pixels while maintaining the original aspect ratio, and the width along the shorter side was then padded black to 1024 pixels, thus matching the image size of our synthetic data. 
The detectors trained with real images only are regarded as baselines, while we add another 5,000 synthetic abnormal images for data augmentation.
The implementation details of different detectors are consistent with the experiments above.
We train and test Comparison Detector on current data splits using the original settings in~\cite{liang2021comparison} for a fair comparison.

Table~\ref{tab:det_public} presents the results for cross-dataset validation.
Our synthesis framework shows the powerful capability of domain generalization in the data augmentation utility, \eg, the baseline with mAP 18.1 \textit{vs.} the augmented model with mAP 20.6 for RetinaNet.
This experimental result indicates that our method can effectively augment various models of downstream tasks even in the face of unseen image domains, making it applicable to various clinical scenarios.
In addition, all the detectors augmented by our synthetic data achieve much higher AP values than Comparison Detector in terms of overall and individual cell types, suggesting that our synthesis-based data augmentation method is more effective in improving model performance.

\section{Conclusion and Discussion}
\label{sec:conclusion}
Cytopathological image synthesis can minimize the efforts in data collection and annotation for cervical abnormality screening. 
However, it is difficult to generate a large-size cytopathological image while synthesizing visually realistic cellular features for small-size abnormal cells.
To overcome this challenge, we propose a novel two-stage cervical cytopathological image synthesis framework based on Stable Diffusion, a foundation model for image synthesis, for data augmentation on cervical abnormality screening. 
The first \textit{Global Image Generation} stage can generate high-resolution normal images full of NILM cells and other natural impurities, which provide visually plausible backgrounds with accurately simulated cell spatial relationships. 
In \textit{Local Cell Editing} stage, a certain number of NILM cells selected from the generated normal images can be converted to abnormal cells of different cervical malignant degrees, thus obtaining synthetic abnormal cytopathological images with box annotations.
The proposed \textit{Normal Image Generator} and \textit{Abnormal Cell Synthesizer} in both stages are built upon pre-trained Stable Diffusion via PEFT techniques for cytopathological image content customization and spatial layout controllability extension, respectively.
The utilization of PEFT techniques significantly reduces the computation budget and the need for large-scale data in training reliable generative models.
Through explicit manipulation of the numbers, the locations, and the cell types of synthetic abnormal cells, large-scale and diverse annotated abnormal images can be created for augmenting cervical abnormality screening in various clinical scenes.
Qualitative and quantitative experiments validate the semantic realism, the flexible controllability, and the data augmentation effectiveness of our proposed synthesis framework.

In the evaluation experiment on the image quality of the synthetic data, we firstly compare our method with different PEFT methods in building \textit{Normal Image Generator} since \textit{Global Image Generation} stage mainly affects the overall realism by creating the majority of the image space.
Qualitative results show that our implementation using LoRA can produce visually realistic normal cervical cytopathological images with semantically plausible cell spatial relationships (\eg, isolated, squeezed, and overlapped) while other methods only generate images of low quality.
Our method also achieves the highest FID scores when using two feature extractors pre-trained on different datasets, which indicates that there is only little discrepancy in feature space between the generated data distribution and the real one in terms of both natural and pathological image domains.
Regarding \textit{Local Cell Editing} stage, we visualize the conversion from NILM cells to abnormal cells of varying malignancy degrees at different locations, demonstrating its flexibility in controlling the synthesis process toward diverse synthetic data creation.
In addition to enriching the training data, the controllability of our synthesis framework enables us to extend it to other potential applications in the future. 
For example, the synthetic data can serve as a comprehensive robustness benchmark~\cite{hu2023label} for rigorously evaluating the performance of the screening systems in detecting abnormal cells at a variety of different locations and cell types. 
It can also augment pathology trainee education~\cite{dolezal2023deep} by reinforcing human understanding of cytopathological manifestations of cervical cancer.

As for the assessment of data augmentation effectiveness, we particularly compare our synthetic data with near-real data, which is created by translating real NIML cells into abnormal cells via our proposed Local Cell Editing. 
Their comparable performances in augmenting different detectors further demonstrate that the generated normal cytopathological images have similar visual appearances to the real ones, further demonstrating the high fidelity of our synthetic data.
Moreover, the proposed synthesis framework is more cost-effective as it can directly synthesize an infinite number of abnormal cervical cytopathological images while the number of near-real data is determined by the real normal images involved in editing.
We also roughly prove the generalization ability of our synthetic data in the data augmentation scenario where the synthetic data is added to the real data from another image domain, \ie, a different cervical cytopathology dataset, to form the training data.
But it is worth mentioning that according to our empirical observation, the growth of the performance will gradually fall into a bottleneck or even become worse if we continue to increase the synthetic data size for data augmentation.
This phenomenon might be attributed to the drastic data distribution shift from the original real data to the synthetic one.
The synthetic data start to dominate the training data distribution when adding too much of them but the testing data still shares a similar distribution with the original real data.
This suggests that a well-designed data augmentation strategy is urgently needed for determining the synthetic data size or even selecting useful synthetic samples in order to achieve effective performance improvement.

As the first work of large-size abnormal cervical cytopathological image synthesis, our study still has several limitations that would be targeted in future works. 
First, our synthesis framework cannot directly produce gigapixel WSIs, which restricts its data augmentation utility to the automatic screening systems that require the context of the entire WSIs.
The expansion of synthetic image size could be further advanced using patch collage algorithms. 
Second, synthesizing a single abnormal image involves multiple steps and models, especially as there are several sampling processes based on diffusion models.
Consequently, our synthesis framework still suffers from low synthesis speed even though the training-free fast sampling algorithm, \ie, DDIM, has already been implemented.
This encourages us to accelerate the sampling during training to achieve rapid and high-fidelity synthesis in the future.
Third, our synthetic data are still not diverse enough to cover a sufficient population distribution since we do not use multi-center training data.
This obstructs the upper limit of its data augmentation effectiveness and restricts its extension to more applications.
The distribution of cervical cytopathological images is very complex in the real world as it contains diverse cell morphology, staining of varying degrees, random impurities, and even contaminations.
It is worth investing more effort and collecting more data to study generative methods for more comprehensive simulation.

\bibliographystyle{IEEEtran}
\bibliography{ref.bib}
\end{document}

%% file: tables/syn_compare.tex
% \usepackage{graphicx}

\begin{table}[]
\centering
\caption{Quantitative comparison between different PEFT methods in building the proposed \textit{Normal Image Generator}. (↓: lower is better). We employ two feature extractors pre-trained on different image domains to compute FID scores.}
\label{tab:syn_compare}
\resizebox{\columnwidth}{!}{%
\begin{tabular}{c|cc}
\hline
Method & FID@ImageNet ↓ & FID@Cytopathology ↓ \\ \hline
Textual Inversion~\cite{gal2022image} & 132.2895 & 114.3035 \\
SVDiff~\cite{Han_2023_ICCV} & 73.2216 & 8.1146 \\
Custom Diffusion~\cite{kumari2023multi} & 46.6576 & 4.6000 \\
Ours & \textbf{43.7446} & \textbf{1.7587} \\ \hline
\end{tabular}%
}
\end{table}

%% file: tables/det_private.tex
\begin{table*}[htbp]
\centering
\caption{Quantitative evaluation of data augmentation effectiveness on our in-house dataset using different synthetic data(↑: higher is better). The best two results are shown in bold and underlined fonts, respectively. 'CD' denotes Custom Diffusion~\cite{kumari2023multi}. 'R', 'S', and 'R2S' refer to 'Real', 'Synthetic', and 'Real-to-Synthetic' images, respectively.}
\label{tab:det_private}
\resizebox{\textwidth}{!}{%
\begin{tabular}{c|c|ccc|cccc}
\hline
 &  &  &  &  & \multicolumn{4}{c}{AP ↑} \\ \cline{6-9} 
\multirow{-2}{*}{Detector} & \multirow{-2}{*}{Training Data} & \multirow{-2}{*}{mAP ↑} & \multirow{-2}{*}{mAP$_{.5}$ ↑} & \multirow{-2}{*}{mAP$_{.75}$ ↑} & ASC-US & ASC-H & LSIL & HSIL \\ \hline
 & R: $3k$ & 14.9 & 27.2 & 15.0 & 13.4 & 21.6 & 20.0 & \textbf{4.5} \\
 & R: $3k$ + S (CD): $5k$ & 15.3 & 26.7 & 15.7 & 14.3 & {\ul 22.4} & 20.4 & {\ul 4.2} \\
 & R: $3k$ + R2S: $5k$ & \textbf{15.7} & \textbf{28.1} & {\ul 16.2} & \textbf{15.2} & 22.1 & {\ul 21.2} & 4.1 \\
 \multirow{-4}{*}{Faster R-CNN} & R: $3k$ + S (Ours): $5k$ & {\ul 15.6} & {\ul 28.0} & \textbf{16.5} & {\ul 14.7} & \textbf{22.5} & \textbf{22.3} & 2.9 \\ \hline
 & R: $3k$ & 15.6 & 28.1 & 16.2 & {\ul 15.4} & 20.2 & 19.8 & \textbf{6.8} \\
 & R: $3k$ + S (CD): $5k$ & {\ul 16.4} & 29.4 & 16.5 & 15.0 & {\ul 23.6} & \textbf{22.2} & 4.7 \\
 & R: $3k$ + R2S: $5k$ & 16.3 & {\ul 29.8} & {\ul 16.6} & 14.5 & 22.6 & 21.9 & {\ul 6.2} \\ 
 \multirow{-4}{*}{RetinaNet} & R: $3k$ + S (Ours): $5k$ & \textbf{16.8} & \textbf{29.9} & \textbf{17.4} & \textbf{16.3} & \textbf{24.4} & {\ul 22.0} & 4.7 \\ \hline
 & R: $3k$ & 15.9 & 28.5 & 16.6 & 16.1 & 20.9 & 21.2 & 5.4 \\
 & R: $3k$ + S (CD): $5k$ & 17.0 & {\ul 30.7} & 17.6 & 17.2 & {\ul 23.6} & {\ul 22.1} & 5.1 \\
 & R: $3k$ + R2S: $5k$ & {\ul 17.3} & 30.3 & {\ul 17.9} & \textbf{18.3} & 22.6 & 21.7 & \textbf{6.5} \\ 
 \multirow{-4}{*}{DINO} & R: $3k$ + S (Ours): $5k$ & \textbf{17.8} & \textbf{31.6} & \textbf{18.1} & {\ul 17.5} & \textbf{25.4} & \textbf{22.2} & {\ul 6.3} \\ \hline
\end{tabular}%
}
\end{table*}

%% file: tables/det_public.tex
\begin{table*}[htbp]
\centering
\caption{Quantitative evaluation of data augmentation effectiveness on Comparison Detector database~\cite{liang2021comparison} using our synthetic data(↑: higher is better). The best results are shown in bold fonts. 'R' and 'S' refer to 'Real' and 'Synthetic' images, respectively.}
\label{tab:det_public}
\resizebox{\textwidth}{!}{%
\begin{tabular}{c|c|ccc|cccc}
\hline
\multirow{2}{*}{Detector} & \multirow{2}{*}{Training Data} & \multirow{2}{*}{mAP ↑} & \multirow{2}{*}{mAP$_{.5}$ ↑} & \multirow{2}{*}{mAP$_{.75}$ ↑} & \multicolumn{4}{c}{AP ↑} \\ \cline{6-9} 
 &  &  &  &  & ASC-US & ASC-H & LSIL & HSIL \\ \hline
Comparison Detector~\cite{liang2021comparison} & R: 4,651 & 17.1 & 31.1 & 13.9 & 14.9 & 4.6 & 25.4 & 21.5 \\ \hline
\multirow{2}{*}{Faster R-CNN} & R: 4,651 & 20.9 & 42.7 & 17.5 & 19.0 & 9.4 & 30.5 & 24.8 \\
 & R: 4,651 + S: 5,000 & \textbf{21.7} & \textbf{43.4} & \textbf{19.8} & \textbf{19.4} & \textbf{11.0} & \textbf{30.8} & \textbf{25.6} \\ \hline
\multirow{2}{*}{RetinaNet} & R: 4,651 & 18.1 & 35.6 & 15.7 & 14.0 & 8.5 & 25.8 & \textbf{24.0} \\
 & R: 4,651 + S: 5,000 & \textbf{20.6} & \textbf{41.3} & \textbf{18.6} & \textbf{18.9} & \textbf{8.9} & \textbf{30.9} & 23.8 \\ \hline
\multirow{2}{*}{DINO} & R: 4,651 & 25.1 & 48.4 & 22.2 & 25.6 & 14.2 & 33.2 & 27.8 \\
 & R: 4,651 + S: 5,000 & \textbf{26.6} & \textbf{48.9} & \textbf{25.2} & \textbf{26.0} & \textbf{15.7} & \textbf{36.5} & \textbf{28.3} \\ \hline
\end{tabular}%
}
\end{table*}